\relax
\documentclass[letterpaper]{article} 
\usepackage{aaai19}  
\usepackage{times}  
\usepackage{helvet}  
\usepackage{courier}  
\usepackage{url}  
\usepackage{graphicx}  
\usepackage{subcaption}
\usepackage{todonotes} 

\usepackage{enotez}
\let\footnote=\endnote

\begin{document}
%
\title{Dungeon Crawl Stone Soup as an Evaluation Domain for Artificial Intelligence}
\author{Dustin Dannenhauer\\
	NRC Research Associate\\
	Naval Research Laboratory\\
	Washington, DC USA\\
	dustin.dannenhauer.ctr@\\~nrl.navy.mil
	\And
	Michael W. Floyd\\
	Knexus Research Corporation\\
	Springfield, VA USA\\
	michael.floyd@\\~knexusresearch.com\\
	\And
	Jonathan Decker\\
	Naval Research Laboratory\\
	Washington, DC USA\\
	jonathan.decker@\\~nrl.navy.mil\\
	\And
	David W. Aha\\
	Naval Research Laboratory\\
	Washington, DC USA\\
	david.aha@\\~nrl.navy.mil}
\maketitle
\begin{abstract}
Dungeon Crawl Stone Soup is a popular, single-player, free and open-source rogue-like video game with a sufficiently complex decision space that makes it an ideal testbed for research in cognitive systems and, more generally, artificial intelligence. This paper describes the properties of Dungeon Crawl Stone Soup that are conducive to evaluating new approaches of AI systems. We also highlight an ongoing effort to build an API for AI researchers in the spirit of recent game APIs such as MALMO, ELF, and the Starcraft II API. Dungeon Crawl Stone Soup's complexity offers significant opportunities for evaluating AI and cognitive systems, including human user studies. In this paper we provide (1) a description of the state space of Dungeon Crawl Stone Soup, (2) a description of the components for our API, and (3) the potential benefits of evaluating AI agents in the Dungeon Crawl Stone Soup video game.
\end{abstract}

\section{Introduction}

Dungeon Crawl Stone Soup\footnote{https://github.com/crawl/crawl} (DCSS) is a single-player, free and open-source rogue-like turn-based video game that consists of a procedurally generated 2-dimensional grid world. To win the game, the player must navigate their character through a series of levels to collect `The Orb of Zot' and then exit the dungeon. Along the way, players encounter a wide variety of monsters and items. Players equip and use items to make themselves stronger or consume them to aid in difficult situations. The world of DCSS is dynamic, stochastic, partially observable, open, and sufficiently complex: the number of states is orders of magnitude larger than games such as Starcraft and Go and the number of instantiated actions the player may take can reach into the hundreds.

DCSS is notoriously hard for humans. Comments such as ``\textit{Wow. I've finally gotten my first win since I started playing, almost exactly 3 years ago.}"\footnote{This comment was posted on October 24, 2018: https://www.reddit.com/r/dcss/comments/9qzfmy/vavp\_mibe\_my \_first\_win\_after\_3\_years/} frequently appear on message boards for DCSS. More experienced players are regularly answering questions and providing advice to newer players. A single game takes on the order of hours to complete; for example, the average playtime for games in a large-scale tournament of human players in 2016 was 8.5 hours.

Rogue-likes are famous for having permanent death: when the player dies, the game ends. Often making a single mistake, or a series of small mistakes will lead to failure. Worse, sometimes these mistakes are only realized hundreds or thousands of turns later. For example, a player may use a one-time-use life-saving item when they could have used a repeatable ability or the player may have trained skills in such a way that they have vulnerabilities against more powerful monsters found later in the game. 

Development of an API for AI agents to play DCSS would offer several desirable aspects for evaluating new and existing AI techniques:

\begin{itemize}
	\item DCSS is a simulated environment which is partially observable, open, dynamic, and stochastic, with an environment model that changes over time (i.e., the probabilities associated with the player's actions change over time)
	\item An environment that requires rich knowledge to play effectively. This includes multiple types of knowledge such as factual knowledge (e.g., the player must obtain 3 runes before entering \textit{The Realm of Zot} level), strategic knowledge (e.g., fighting a hydra monster with a non-fire bladed weapon should be avoided), and descriptive knowledge (almost every aspect of the game comes with an English text description designed for a human user - this includes all objects, tiles, and monsters).
	\item A game that requires long-term strategic planning where early decisions can have a significant impact on later game play. Poor decisions early can be irreversible and have significant consequences (e.g., permanent death). 
	\item An environment that does not penalize slow reaction times. DCSS is a turn based game with no time limit on deciding which action to take next. New players are often advised to pause when they realize they are in a dangerous situation in order to (1) carefully consider all of their options and (2) learn about the monsters and items in the current situation from online knowledge bases including a wiki, forum, and live IRC chat with other players.
	\item There is existing data on human performance for thousands of previous games played. This opens the possibility for comparison between human and intelligent agents using DCSS.
	\item A game interface that enables multiple spectators to watch the player and interact with them via natural language text dialog. This provides multiple opportunities for human-agent interaction, such as explanation, human-agent teaming, and intelligent tutoring.
\end{itemize}

In the following section, we describe the state space and environment properties that make DCSS an interesting research domain. Following that, we describe the skill level of human DCSS players from an annual tournament. Next we highlight current efforts to build an API for DCSS followed by a  description of cognitive systems and AI approaches that could be evaluated using such an API. We then discuss related work and conclude with a summary and next steps.

\section{DCSS State Space and Environment Properties}

We now identify in more detail properties of DCSS that lead to its high complexity, followed by a lower bound theoretical analysis of the state space and action space.

\begin{itemize}
	\item 650+ unique monster types which the player may encounter, many of which require specific actions, attributes, or special knowledge to be able to defeat. For example, if you attack a hydra monster with a weapon that has a blade (e.g., axe, sword) you will chop off it's head and it will grow more in its place, and as a result become much stronger. A good approach to defeating hydras is to have a bladed weapon enchanted with fire (which sears the wound) or use a blunt force object such as a mace. Magic also works.  
	\item 13,800 possible starting character configurations formed by choosing: one of 23 species (e.g., vampire, ogre), one of 24 backgrounds (e.g., fighter, wizard, berserker), and one of 25 deities for your character to worship that may provide additional benefits (e.g., worshiping \textit{Gozag Ym Sagoz} turns slain enemy corpses into gold). Some are considered easier than others; a minotaur berserker worshiping Trog is the recommended starting character for new players who have yet to win the game.
	\item 31 skills (e.g., fighting, short blades, hexes, charms, shields) and 3 attributes (strength, intellect, dexterity) that are increased by spending experience points. The value of each skill ranges from 0 to a maximum of value of 27. Spending experience is permanent and cannot be undone. Poor decisions in allocating experience points for skills and attributes is a major cause for players not being able to win the game, since improperly raising your attributes leads to deficiencies against certain monster types later on. It is also specific to the items and spells a character will focus on, which often changes during the course of a game. Finding a rare item meant for melee may warrant an entire strategy change for a character that is currently magic-based. It is not always an easy decision because there may not be enough time to raise skill and attribute values before encountering monsters which require such skills to be defeated. 
	\item 100+ spell actions a player can learn. A player can only know a maximum of 21 spells at any given time. Spells have unique effects that sometimes require careful planning. Some spells buff the player with attributes that affect later actions. For example, when in a situation where time is of utmost importance, often casting a spell to temporarily increase the speed of the player should be done first.
	\item 48 unique types of melee and ranged weapons that player may encounter and use. Each weapon may be branded to give it additional effects (e.g., fire, frost, venom) that may do additional damage and cause special effects (e.g., a monster hit with a venom brand will gain a temporary poison status that deals damage over time).
	\item 15 runes to be collected. Runes are special items that do not take up inventory space and serve to enable the player to visit new branches (series of levels) of the dungeon. Collecting a minimum of three runes is necessary to access the \textit{Realm of Zot} level, which is required to win the game. While 3 runes are the minimum requirement, many players challenge themselves to see how many runes they can acquire. Runes are associated with special areas in the game (i.e. the serpentine rune requires fighting snake-themed monsters and a resistance to poison is highly recommended). Some runes are significantly more difficult to obtain than others.
	\item Approximately 65,000 to 80,000 turns is typical for a 3-rune game. Turns can be considered an approximation of the number of actions taken. This can vary depending on the speed of the player, which may be faster or slower than the turn rate, in which case a fast player may take 2 actions in 1.5 turns or a slow player may take 1 action in 1.5 turns. Speed is an attribute of the player's character depending on their attributes an items (e.g. equipping heavier armor can slow attack speed; other items may increase or decrease the player's movement speed). Speed here does not refer to how long the player takes in deciding the next action to take. 
	\item 40+ consumable resource-based items including: 18 potions, 10 scrolls, 11 wands and a small number of specialty items. Potions and scrolls are single-use and offer some of the most important life saving capabilities, such as a scroll of blinking which instantly teleports the character to another tile within line of sight of the player.
	\item Players may encounter more than 70,000 tiles before completing a game. A tile is a location on the grid that may hold a combination of monsters, items, and special terrain features (e.g., lava, water, steam).
	\item 100+ levels. Levels are composed of tiles that are procedurally generated to form rooms, passageways, etc., using a variety of terrain elements such as walls, shallow water, deep water, lava, etc. Levels are connected via staircases that act as portals from one level to the next. A 3-rune game requires visiting at least 45 levels. Most levels have a static arrangement of tiles except for two special levels, Abyss and Labyrinth, where the number of tiles is infinite and the layout of tiles outside the player's line of sight constantly changes.
	\item Partially observable: the player does not see a tile until it is within line of sight, which is normally within seven tiles of the player in any cardinal direction.
\end{itemize}

\begin{figure}[t]
	\centering
	\includegraphics[scale=0.33]{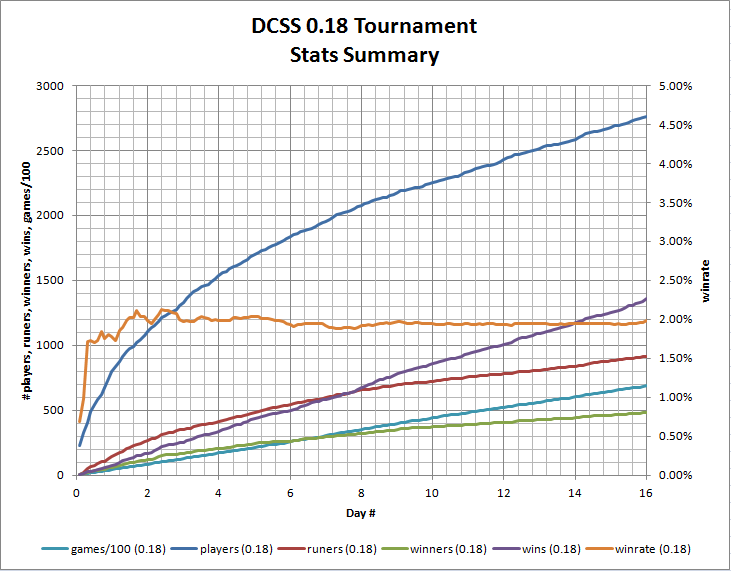}
	\caption{Results from the v. 0.18 tournament}
	\label{tournyresults}
\end{figure}

\begin{itemize}
	\item Dynamic: monsters take their own actions independent of the player, and there are time-based events such as entrances to special areas that close after a time limit (e.g., volcano and sewer levels).
	\item Stochastic: most actions (e.g., melee attacks, spells) are probabilistic and often fail.
	\item Natural language text accompanies every item and action in the game. The player can ask for a description of any tile, object, monster, etc., within view or in the player's inventory.
	\item Permanent death: if the player dies, the game ends and they must start again in a newly generated world. The only way to replay a game is to manually set the seed for the procedural generation.
\end{itemize} 

\begin{figure*}[h]
	\centering
	\begin{subfigure}[b]{0.5\textwidth}
		\centering
		\includegraphics[height=2.5in]{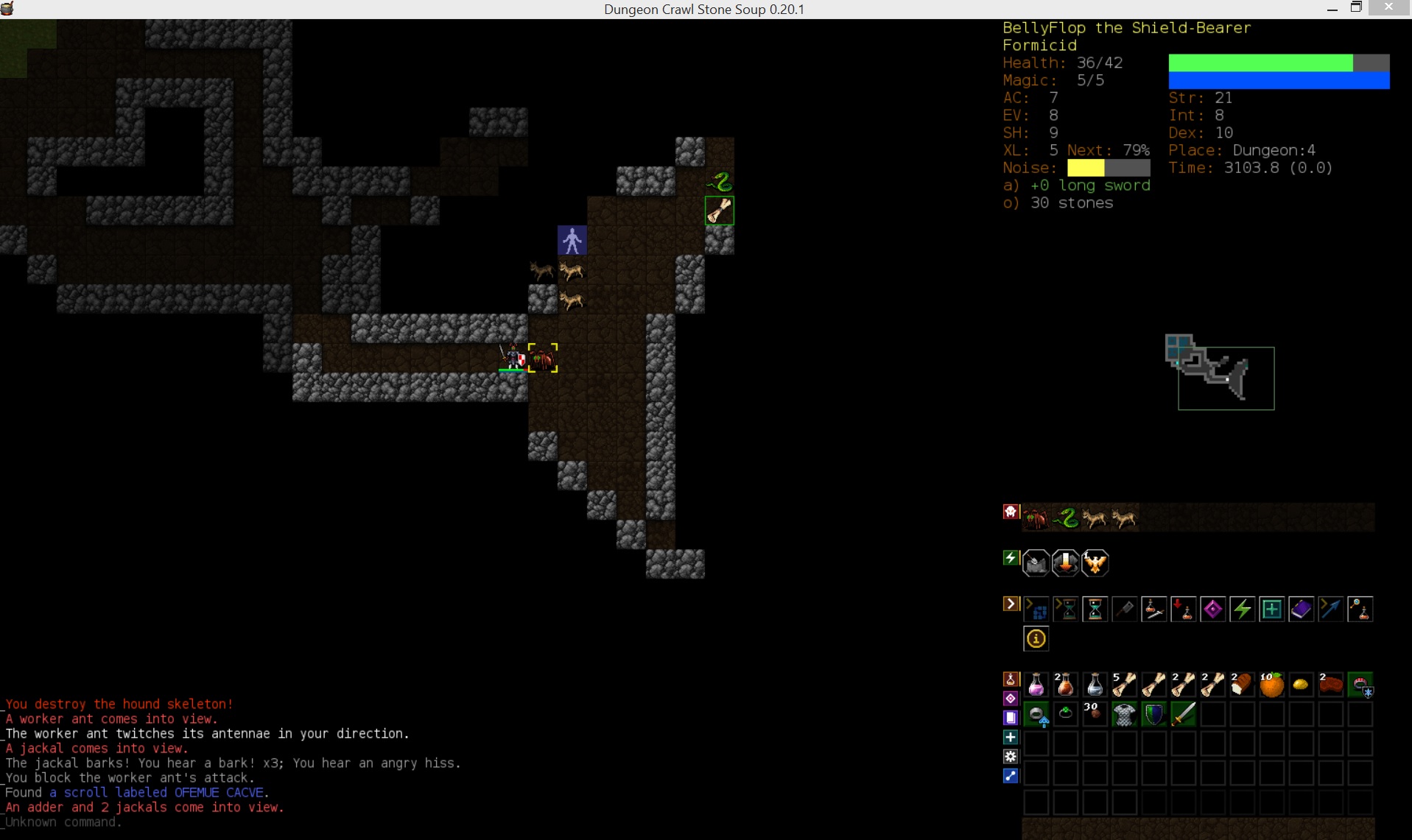}
		\caption{Screenshot of the full DCSS Game in Progress}
		\label{full-game-screenshot}
	\end{subfigure}		
\hspace{1cm}
	\begin{subfigure}[b]{0.4\textwidth}
		\centering
		\includegraphics[height=1in]{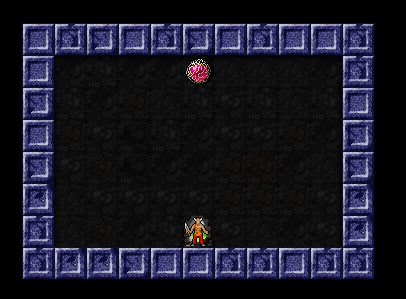}
		\caption{Open-room Custom Scenario}
		\label{large-custom-scenario-screenshot}
		
		\vspace{2ex}
		
		\centering
		\includegraphics[height=1in]{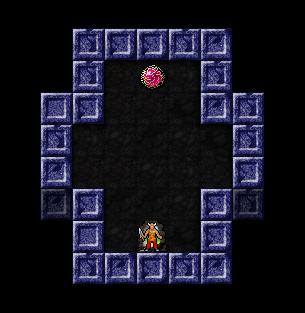}
		\caption{Small Custom Scenario}
		\label{small-custom-scenario-screenshot}
	\end{subfigure}
	\caption{Screenshots of Dungeon Crawl Stone Soup}
	\label{all-screenshots}
\end{figure*}

We now give a lower bound complexity analysis of the state space for a complete game using the following assumptions. These assumptions are lower bounds on the numbers of tiles, items, monsters, etc the player would encounter in a 3-rune game. Technically DCSS has an infinite state space.

\begin{itemize}
	\item 70,000 tiles
	\item 900 items (estimated 20 items per level, 45 levels)
	\item 2000 monsters
\end{itemize}

For simplicity, let's assume that monsters and items will not be generated on the same tile. With these minimum assumptions, the state space is $$ |S| = 70000^{2900} \approx 10^{14000}$$ which is significantly more than Starcraft (estimated at lower bound of $10^{1685}$), Go (estimated at around $10^{170})$, and Chess (estimated at $10^{50}$) \cite{ontanon2013survey}. Starcraft however has a significantly higher action space estimated at $10^8$ \cite{vinyals2017starcraft} where as we estimate the number of grounded actions for DCSS to be no more than 1000 in any given state. A primary difference between Starcraft and DCSS is real-time decision making. Since DCSS does not penalize long reaction times, cognitive approaches for more deliberate reasoning (such as planning and inference mechanisms) can be effectively evaluated in DCSS, while still ensuring a highly complex environment.

\section{Annual DCSS Tournament}

With every major release of the game (e.g., v. 0.17, v. 0.18, v. 0.19) there is a tournament held with thousands of players that spans 16 days. During this time, players try to collect as many points as possible by playing a variety of different character configurations (i.e.,  species, background, and deity combinations). The results for the tournament using v. 0.18 of the game \cite{dcssV018TournResults} were posted on June 3rd, 2016 and some of these results are shown in Figure \ref{tournyresults}. Tournaments attract the best human players to show off their skill and serves as one possible benchmark with which to evaluate AI agents against humans. Especially for those unfamiliar with DCSS, here are some stats to give a better idea of what is required by human players: 

\begin{itemize}
	\item The average won game took around eight and half hours of human playtime.
	\item The fastest run in the v. 0.18 tournament is 41:00 minutes.
	\item 2500 human players competed in the tournament, with only about 500 players winning a game.
	\item The overall win rate of games attempted is slightly more than 2\% by the end of the tournament.
\end{itemize}

There are many intermediate metrics that can be used to measure AI performance besides winning the game and the score the player has accumulated. These could include number of runes collected, number of levels reached, time, number of actions the agent has taken, number of monsters killed, etc. Our API will maintain a history of the players actions and other data to output metrics such as these.

\section{An API for AI Agents for DCSS}
Previous efforts to build a bot for DCSS (see Related Work) used hand coded expert knowledge. The focus of this API is to provide a game state data object to a computer program (the AI) and accept action commands which will then be executed. The API we are building has the following properties:

\begin{itemize}
	\item Ability to send actions and receive game state from a webserver client. This allows observers (including humans) to watch and interact with the agent via chat messages. 
	\item Game state data objects in a structured representation format like PDDL \cite{mcdermott1998pddl}
	\item A PDDL domain file containing models for non-combat actions
	\item Support for treating multi-step actions as single commands. Often a user must issue multiple keypresses to take certain actions such as first choosing the throw command and then choosing a target. Our API allows an agent to issue a single command and takes care of the execution of all the steps involved to send to the game engine.
	\item Ability to run agents in custom levels. Figures \ref{large-custom-scenario-screenshot} and \ref{small-custom-scenario-screenshot}  show custom scenarios that are simply empty rooms with no monsters or objects except for a single orb. Items and monsters are easily added when designing the scenario.
\end{itemize}

\section{Promising Research Questions Facilitated by the DCSS Domain}

Dungeon Crawl Stone Soup is unique in that it is a highly complex game both in terms of state and action space while also being difficult for humans. Playing well requires large amounts of different types of knowledge (factual knowledge vs. strategic knowledge for example). Human performance data is available to compare against AI systems. Additionally, almost everything in the game is accompanied by natural language text. Because of these qualities, DCSS is an excellent research testbed to explore solutions to the following problems:

\begin{itemize}
	\item \textbf{Achievement goals vs. learning goals:} An agent may find it is constantly dying in a situation and should consider querying the online wiki or ask a question on an online IRC chatroom to understand why it's failing. Once it has found knowledge relevant to the problem the agent must decide how to use such knowledge. The topic of achievement goals vs. learning goals is an open problem in goal reasoning.
	\item \textbf{Planning and acting with learned models:} The probabilities of an agent's effects (e.g., combat, likelihood to land a hit, likelihood to block or dodge an attack) change as the agent gets stronger and with respect to different types of monsters. How can an agent plan and act with these changing models, and how can it update it's models?
	\item \textbf{Intelligent assistants and tutors:} Could you develop an assistant that aids players in completing the game by offering advice and/or guidance? Perhaps an intelligent tutoring agent could observe a human player fail repeatedly in a situation (i.e. every time the human player faces a hydra monster, the character dies) and generate custom scenarios designed to teach the human player proper strategies to running from or defeating hydras. This could include lessons in allocating skill points, selecting among a variety of weapons, and using a variety of escape related items.
	\item \textbf{Explainable planning and goal reasoning agents:} The interpretability of AI systems has been a popular topic of workshops and related events since 2016, and in 2017 DARPA launched the Explainable AI (XAI) Program. Most of these efforts have focused on providing  transparency to the decision making of machine learning (ML) systems in general, and deep networks more specifically\footnote{Exceptions, for example, include the broader intent of the Workshops on XAI at IJCAI-17 and IJCAI-18}. While XAI research on data-driven ML is well-motivated, AI Planning is well placed to address the challenges of transparency and explainability in a broad range of interactive AI systems. For example, research on Explainable Planning has focused on helping humans to understand a plan produced by the planner (e.g., \cite{Sohrabi11,Bidot10}), on reconciling the models of agents and humans (e.g., \cite{chakraborti17}), and on explaining why a particular action was chosen by a planner rather than a different one (e.g., \cite{smith12,langley17,fox2017explainable}.
	
	DCSS has a sufficiently rich environment containing different types of knowledge that make understanding decision making difficult. Novice players watching an expert player may not understand why decisions were made or actions were taken. Thus, DCSS could be a suitable environment for evaluating agents that explain their planning and other decision making components to humans.
	
	\item \textbf{Knowledge Extraction from Games:} Dungeon Crawl Stone Soup is a knowledge rich game which takes humans many hours of playing and reading before acquiring enough knowledge to complete the game. While our API provides a starting point to use techniques such as automated planning, there are opportunities for new approaches to knowledge extraction that could be evaluated with DCSS.
	\item \textbf{Single-Policy Reinforcement Learning (RL) vs. Hierarchical or Component-Based Approaches}: An open question is whether RL approaches can learn a single policy for large state action spaces or whether policies per individual goals or tasks scale better. Does having an explicit goal representation (for example, the agent may have the goal of a nearby hydra to be dead: \textbf{dead(hydra)}) and goal-specific plans or policies lead to more manageable decision making rather than relying only on knowing the best action to take in any given state to reach some reward?
	\item \textbf{Curriculum-based RL:} In environments such as DCSS there is delayed reward. The most obvious reward function is winning a game but since this requires tens of thousands of actions to do so, intermediate reward functions will be needed. The player's cumulative game score could be used, but this may not be enough to determine such actions as spending experience points to increase skill levels. Could an agent identify for itself what rewards it should pursue? Will a curriculum-based RL approach lead to an agent that can complete the game?
	\item \textbf{Execution monitoring, replanning and goal reasoning:}
Consider an example where an agent is executing a plan to achieve the goal of killing a monster and the agent observes a rare weapon item nearby. The agent may decide to replan in order to pick up the object and use it to kill the monster, but to do so would require kiting the monster around an obstacle to reach the item without being attacked first. Can we build agents capable of reasoning about goals and plans in an environment such as DCSS that could lead to such behavior?
\end{itemize}

\section{Related Work}

Surprisingly, little work in AI has made use of rogue-like video games. \citeauthor{steinkraus2004combining} (\citeyear{steinkraus2004combining}) used an extremely simplified version of NetHack\footnote{https://www.nethack.org/} to evaluate learning abstractions for large MDPs. More recently \citeauthor{winder2018keg} (\citeyear{winder2018keg}) identified NetHack as  ``an immensely rich domain'' worth using to evaluate concept-aware task transfer as future work. Dungeon Crawl Stone Soup is a richer domain than NetHack in many aspects (e.g., number of spells, number of starting characters), although an API for NetHack would also be a contribution to the AI community.

Computer programs to play DCSS and NetHack have been handcoded. \textbf{qw}\footnote{https://github.com/elliptic/qw} is the best known bot for DCSS achieving the highest winrate of about 15\% for 3-rune games with the starting character of Deep Dwarf Fighter worshipping Makhleb, and also achieves a 1\% winrate for a 15-rune game with a Gargoyle Fighter worshipping Okawaru. The first bot to beat NetHack with no human intervention was created by Reddit user \textit{duke-nh}\footnote{https://www.reddit.com/r/nethack/comments/2tluxv/ ~~~~yaap\_fullauto\_bot\_ascension\_bothack/}. Both of these bots rely extensively on expert-coded knowledge and rules, and do not perform learning. They demonstrate that programs are capable of beating these games and being open source, provide baselines for AI agents playing these games. 

Video games such as DCSS offer some of the complexities of real-world environments: dynamic, partially observable, open, etc... in a software simulation that is often cheaper and/or faster to evaluate new approaches. A number of simulated environments have released over the last few years: the MALMO API for Minecraft from Microsoft Research \cite{johnson2016malmo}, the Starcraft II API \cite{vinyals2017starcraft} from Deepmind and Blizzard, and the ELF platform for Game Research \cite{tian2017elf} from Facebook. Dungeon Crawl Stone Soup (DCSS) fills a needed gap in the available simulated environments because it offers high complexity, partial observability, and non-determinism, yet without the difficulty of decision-making in real-time. This makes DCSS more manageable for agents that may require deliberation in their decision making such as automated planning, inference, and online learning mechanisms.

\section{Conclusion}

DCSS is an excellent evaluation testbed for many problems in artificial intelligence and cognitive systems, and is supported by an active community of players and developers. We describe properties of DCSS that warrant it's consideration as an evaluation testbed, particularly because it is partially observable, dynamic, stochastic, open, is surrounded by an active community of players and developers, and requires a variety of decision making capabilities in order to win the game. We include a theoretical lower bound of the state and action space complexity analysis, showing it has more states than Starcraft, Go, and Chess. We also described an ongoing effort to build an API for AI agents to play and be evaluated in this game. Future work on the API will be to add support for combat actions, a feature vector representation of the state, and provide planning and reinforcement learning agent tutorials. 

\printendnotes
\clearpage


\bibliographystyle{named}
\bibliography{references}


\end{document}